# Biogeography-Based Optimization of reinforced concrete structures including static soil-structure interaction

Iván A. Negrin[1*], Dirk Roose[2a], Ernesto L. Chagoyén[1b], Geert Lombaert[3c]

[1]*Department of Civil Engineering, Faculty of Construction, Marta Abreu Central University, Santa Clara, Cuba.*
[2]*Department of Computer Science, KU Leuven, Belgium.*
[3]*Department of Civil Engineering, KU Leuven, Belgium.*

**Abstract.**   A method to minimize the cost of the structural design of reinforced concrete structures using Biogeography-Based Optimization, an evolutionary algorithm, is presented. SAP2000 is used as computational engine, taking into account modelling aspects such as static soil-structure interaction (SSSI). The optimization problem is formulated to properly reflect an actual design problem, limiting e.g. the size of reinforcement bars to commercially available sections. Strategies to reduce the computational cost of the optimization procedure are proposed and an extensive parameter tuning was performed. The resulting tuned optimization algorithm allows to reduce the direct cost of the construction of a particular structure project with 21% compared to a design based on traditional criteria. We also evaluate the effect on the cost of the superstructure when SSSI is taken into account.

## 1. Introduction

The optimization of structures is very relevant in the context of structural design, in order to minimize construction costs and material use, amongst others in view of the corresponding environmental impact. The optimization of reinforced concrete (RC) structures is more complex than of other types of structures such as steel structures, due to some characteristics, e.g. the anisotropy of reinforced concrete as a building material and its effect on the structural behavior.

Many authors have studied the optimization or RC structures. To reduce the complexity of the problem, several simplifications are made in modelling. The simplest approach is to optimize isolated elements. Malasri *et al.* (1994) use a Genetic Algorithm (GA) to optimize the bending capacity of a simple beam. Yeo and Gabbai (2011) optimize the design of a beam with two independent objectives: cost and embodied energy. The resulting designs for these two goals are compared. Many other simplifications are made in modelling, including the use of calculation area of reinforcement. Medeiros and Kripka (2016) use Harmony Search (HS) and Simulated Annealing (SA) for the discrete optimization of the design of RC columns subjected to uniaxial flexural compression using economic objective. Afshari *et al.* (2019) perform multi-objective optimization of the design of a RC beam, with cost and deflection objectives. A review of some multi-objective optimization algorithms is made, including some metaheuristic algorithms such as GA, SA, HS or Particle Swarm Optimization. Luévanos-Rojas *et al.* (2020) optimize the design of rectangular cross-section beams with straight haunches with a minimum cost objective. This approach of analyzing simple elements does not take into account the redistribution of forces in statically indeterminate structural assemblies, where changes in each element also affect the elements connected to it, and where the objective function reflects this interaction.

Others have focused on case studies considering plane frame structures. Yeo and Potra (2015) perform continuous optimization of the design of a RC single frame (one beam, two columns) with economic and $CO_2$ footprint objectives. Serpik *et al.* (2016) use GA in discrete optimization of a similar structure with an economic objective. The physically nonlinear behavior of concrete and reinforcement, as well as the possibility of crack formation in concrete are taken into consideration. Guerra and Kiousis (2006) optimize the design of multi-bay and multi-story RC frame structures, obtaining savings up to 23% over typical design methods. Kripka *et al.* (2015) use HS to optimize the design of multi-bay RC frames structures. The dimensions and concrete strength are represented by discrete variables; the diameters of the reinforcement bars of columns are limited to commercially available ones; the beam's steel areas are considered as continuous. Triches *et al.* (2019)

---

*Corresponding author, Lecturer, Ph.D. Student
 E-mail: indiaz@uclv.cu
[a]Ph.D., Professor
 E-mail: Dirk.Roose@kuleuven.be
[b]Ph.D., Professor
 E-mail: chagoyen@uclv.edu.cu
[c]Ph.D., Professor
 E-mail: Geert.Lombaert@kuleuven.be

also use HS in the design optimization of similar structures, considering the automated grouping of columns. These other considerations are closer to practice, although the three-dimensional nature of the structures and the design forces are not taken into account.

Another generally ignored modeling aspect in structural optimization problems is soil-structure interaction (SSI). Authors usually assume structures with idealized support conditions (fixed or pinned), but the assembly of soil and foundation is not perfectly rigid: support displacements (settlements) produce an internal force redistribution in the entire superstructure, i.e. SSI affects the responses of the whole system during loading-unloading processes.

Depending on the load features taking into account (static or dynamic) a distinction can be made between *Static*-SSI (SSSI) and *Dynamic*-SSI, as in Khatibinia *et al.* (2013), in which the soil is assumed to be layered with constant material properties along its depth, and the foundation is modelled as a rigid strip footing. Plane strain conditions with a constant soil thickness corresponding to the inter-frame distance are considered. A modified pressure-independent multi-yield surface J2 plasticity model Zhang *et al.* (2008) is adopted as constitutive model of soil. The shear wave velocity $V_s$ and the friction angle $\varphi$ are considered as parameters of the soil layers.

Unlike, in the present study, SSSI is integrated into the process, adapting a Winkler model, as proposed by Klepikov *et al.* (1987), to relate the contact soil pressure $p$, to the foundation settlement $S$, taking into account linear and non-linear soil behavior with only one equation (Chagoyen *et al.* 2018, Negrin *et al.* 2019a).

Some of the features considered in the present study are: constructive dimensioning of the elements; real (and not calculation) areas for longitudinal/transversal (shear) reinforcing steel; using commercially available bar diameters; bars cutoff (detailing) and its real distribution within the direct cost calculation used to define the objective function (see Fig. 1) (Negrin *et al.* 2019a, b). This increases the complexity of the mathematical formulation of the problem.

In some of the cited works, the authors create their own calculation engines to perform the structural modeling, analysis and design. This limits the selection of case studies, leading to the need for simplifications. In this paper, we use the CSi-SAP2000 platform as calculation engine, via the Applied Program Interface (API) SAP2000-MATLAB. This makes it possible to consider complex case studies and to take into account several factors that allow us to obtain results that are relevant from a practical structural engineering point of view. A drawback of the corresponding increased complexity of the optimization problem is the computational cost. Hence, designing a good optimization strategy is essential.

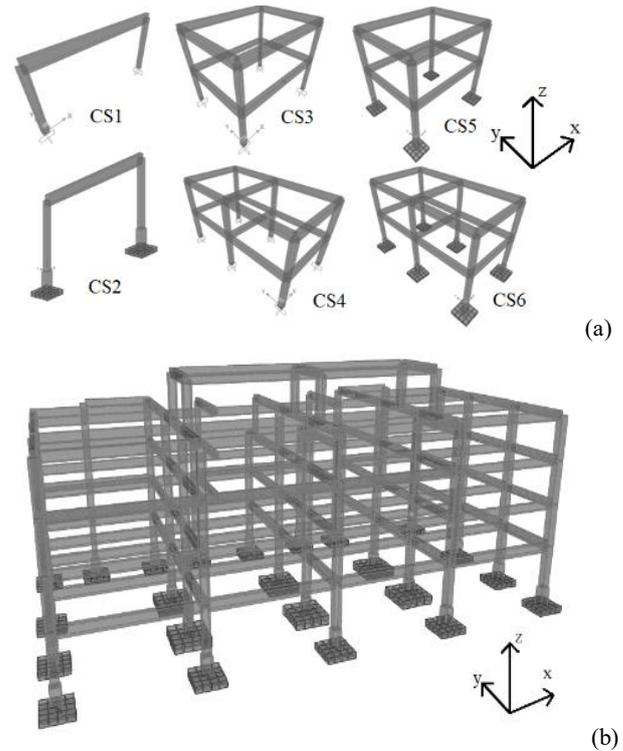

Fig. 2 Case studies used: (a) six simple cases, (b) Síndico House project

The proposed optimization methodology is tested with six models (Fig. 2(a)), representing three structures with and without SSSI included. These simple case studies allow, in a relatively straightforward way, to carry out studies and to draw conclusions on the optimization of RC structures and on the performance of the optimization methods. Table 1 summarizes the number of variables in each case, as well as the number of possible solutions, taking into account the number of values that each (discrete) variable can take.

The structure of the paper is as follows. In section 2 we formulate the optimization problem and explain the various terms in the objective function. In section 3 we review the main properties of Biogeography-Based Optimization and in section 4 we propose a methodology for parameter tuning using a utility derived from the average performance curve. In section 5, we apply the tuned method to the case studies, and also to a more complex model, the Síndico House Project, shown in Fig. 2(b).

## 2. Formulation of the optimization problem

The formulation of the optimization problem is an essential part of the process. Herein, a formulation similar to that used in (Negrin *et al.* 2019a, b) is adopted. The essential aspects are briefly explained below.

### *2.1 Objective function*

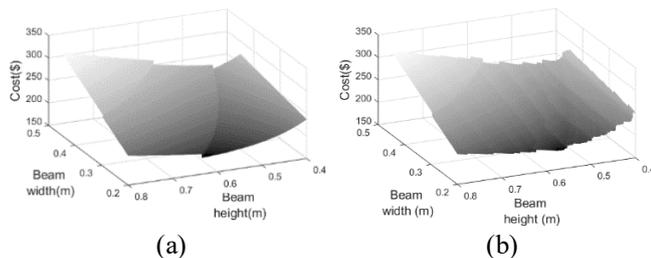

(a)                  (b)

Fig. 1 Differences in the response surface for direct cost for a simply supported RC beam under distributed load: (a) using calculation area, (b) using real area

Table 1 Summary of the six case studies used

|  | CS1 | | CS2 | | CS3 | | CS4 | | CS5 | | CS6 | |
|---|---|---|---|---|---|---|---|---|---|---|---|---|
|  | Groups* | # var | Groups* | # var | Groups* | # var | Groups* | # var | Groups* | # var | Groups* | # var |
| Beams | 1 | 2** | 1 | 2** | 2 | 4 | 1 | 2 | 2 | 4 | 1 | 2 |
| Columns | 1 | 2** | 1 | 2** | 1 | 2 | 2 | 4 | 1 | 2 | 2 | 4 |
| Foundations | - | - | 1 | 1 | - | - | - | - | 1 | 1 | 2 | 2 |
| Possible solutions | 324 | | 1350 | | 2430 | | 1215 | | 21870 | | 98415 | |

* Elements that take same variable values are considered a group, e.g. beams in "x" axis direction
** The total number of variables is 3 in CS1 and 4 in CS2, since the width of the beam and columns is a single (common) variable

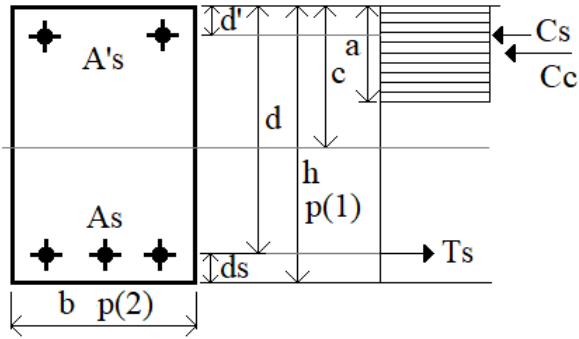

Fig. 3 Rectangular beam cross-section and diagram of forces, including some terms, $p(1)$ and $p(2)$ are variables (dimensions), $d$ is the effective depth, $As$ and $A's$ are the steel area in traction and compression, $d'$ and $ds$ are distances from extreme fibers to the centroid of longitudinal reinforcement in traction and compression, $a$ is the depth of equivalent rectangular stress block.

Since the cost reduction is the goal in this study, the objective function is the total direct cost of construction of the structure (beams, columns and foundations), formulated according to PRECONS (2008) as in Eq. (1),

$$F = C_{Beams} + C_{Col} + C_{Found} \quad (1)$$

where the costs of the beams, columns, and foundation are elaborated as follows.

$$C_{Beams} = C_{FB} + C_{StElB} + C_{StAsPlB} + C_{RBElB} + C_{RBAsPlB} + C_{CoElB} + C_{CoPlB} \quad (2)$$

$$C_{Columns} = C_{FC} + C_{StElC} + C_{StAsPlC} + C_{RBElC} + C_{RBAsPlC} + C_{CoElC} + C_{CoPlC} \quad (3)$$

$$C_{Found} = C_{Exc} + C_{FF} + C_{StElF} + C_{StAsPlF} + C_{RBElF} + C_{RBAsPlF} + C_{CoElF} + C_{CoPlF} + C_{Rf} \quad (4)$$

In Eq. (2)-(4), $C_{F*}$ is the cost of formwork, $C_{StEl*}$ is the cost of the stirrups elaboration, $C_{StAsPl*}$ is the cost of the assembly and placement of stirrups, $C_{RBEl*}$ is the cost of longitudinal reinforcement bars elaboration, $C_{RBAsPl*}$ is the cost of longitudinal reinforcement bars assembly and placement, $C_{CoEl*}$ is the cost of concrete elaboration and $C_{CoPl*}$ is the cost of concrete placement. In Eq. (4), $C_{Exc}$ is the excavation cost and $C_{Rf}$ is the refill cost. All of these direct construction costs are obtained by multiplying the volume of work with the corresponding unit cost in PRECONS (2008). Two examples can be seen in Eq. (5)-(6).

$$C_{CoElB} = \left(\sum_{i=1}^{N_b} h_i \cdot b_i \cdot L_i\right) C_{UnCEl} \quad (5)$$

$$C_{exc} = \left(\sum_{i=1}^{N_f} V_{exc}\right) C_{UnExc} \quad (6)$$

In Eq. (5), $N_b$ is the number of beams, $h_i$, $b_i$ and $L_i$ are respectively the cross section dimensions (see Fig. 3) and the length of beam $i$ and $C_{UnCEl}$ is the unit cost of concrete elaboration (\$/m³). In eq. (6), $N_f$ is the number of foundations, $V_{exc}$ is the volume of excavation of one foundation (see Fig. 4) and $C_{UnExc}$ is the unit cost of excavation (\$/m³).

### 2.2 Design Variables

As mentioned, all design variables are discrete and can only take values that occur in practice. For the beams and columns, each group has two design variables, which specify the dimensions of the cross sections (see Fig. 3), limiting possible values to multiples of 5 cm for reasons of buildability. For the foundations, the design variables are the rectangularity (ratio of the sizes along the x-axis ($L$) and the y-axis ($B$)) (see Fig. 4). For the type of concrete, the fundamental variable is the specified compressive strength of concrete ($f'c$); other parameters (E, etc.) depend on $f'c$.

### 2.3 Constraints

Constraints are all conditions that must be satisfied by the design variables to ensure that a design is feasible. They are defined mathematically as specific lower, upper or equality limitations imposed on the design variables or on the assigned parameters.

There are two types of constraints. *Design (explicit) constraints* are those that are imposed on the design variables directly and appear for various reasons, such as functionality, manufacturing, transport or esthetic, and can be presented as: $X_{min} \leq X \leq X_{max}$. *Behavioral (implicit) constraints*, which are sometimes called state equations, are indirect. In the context of the present work, these constraints deal with the fulfilment of the limit states, i.e., they define the values that the variable parameters must meet to satisfy the behavioral requirements.

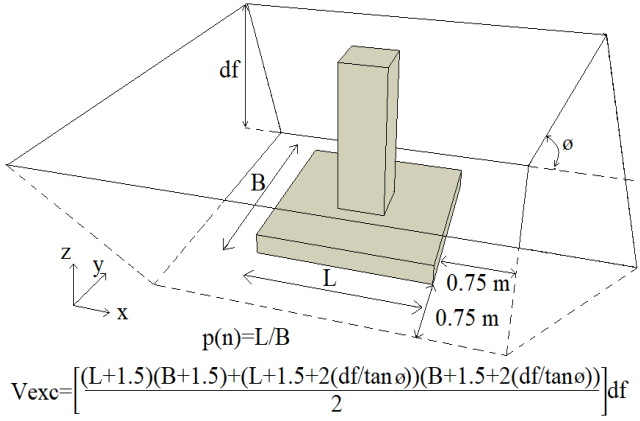

Fig. 4 Typical shallow foundation and excavation scheme. $p(n)$ is the rectangularity (variable), $V_{exc}$ is the volume of excavation of one foundation

In structural optimization, the behavioral constraints are usually set by design standards, for example, maximum allowable stresses/displacements, crack width, deflections. Ultimately, these state equations govern the design itself and are more complex than the other constraints (Negrin 2009).

Many constraints must be considered in the present optimization problem. Examples of most relevant state equations for the design of RC beams are the following strength criteria:

$$Mu \leq \phi Mn = \phi \left[ 0.85 f'c \cdot a \cdot b \left( d - \frac{a}{2} \right) + fy \cdot A's (d - d') \right] \quad (7)$$

$$Vu \leq \phi (Vs + Vc) = \phi \left[ \frac{A_v \cdot fy_t \cdot d}{s} + 170 \sqrt{f'c} b_w \cdot d \right] \quad (8)$$

In Eq. 8, $A_v$ is the area of shear reinforcement within spacing $s$, in m² and $b_w$ is $b$ in rectangular cross sections. The relevant strength (or limit state) constraints for the design of RC beams are automatically satisfied through the API SAP2000-MATLAB platform, which computes the structural design according to the standards. The stiffness (or serviceability) constraints, which limit the beam deflections ($\Delta \leq (L/180)$), are checked and if not satisfied, the objective function is penalized. The same strategy is followed when other constraints associated with the distribution of longitudinal reinforcement, for example, are not met, i.e., *constructive constraints*.

For the design of RC columns, the following strength constraints must be satisfied (Eq. (9)-(10)).

$$Pu \leq \phi Pn = \phi \left[ 0.85 f'c \cdot b \cdot a + A's \cdot fy - As \cdot fy \right] \quad (9)$$

$$Mu \leq \phi Mn = \phi \left[ 0.85 f'c \cdot b \cdot a \left( \frac{h}{2} - \frac{a}{2} \right) + A'sf's \left( \frac{h}{2} - d' \right) + Asfs \left( \frac{h}{2} - d_s \right) \right] \quad (10)$$

Related to the serviceability limit state, another relevant constraint is the limit on the top building displacement (in wind loading) as shown in Eq. (11). $H$ is the total height of the structure.

$$D_{top} \leq \left[ \frac{H}{450} \right] \quad (11)$$

Examples of the most relevant state equations for shallow foundations, for the geotechnical design by I Limit State (ONN (NC) 2014), are:

$$SF_{overturn.} = \frac{M_{stabil}}{M_{turn\,over}} \geq 1.5 \quad (12)$$

$$H^* \leq 0.75 \cdot b' \cdot l' \cdot c^* + N_c^* \cdot tan\phi^* \quad (13)$$

$$N_c^* \leq Q_{bt}^* \quad (14)$$

which express in that order the requirement of an adequate safety factor against overturning, and the sliding and strength condition that must be satisfied in limit state design.

For geotechnical design by the II Limit State (ONN (NC) 2014), the following constraints hold:

$$SF_{overturn.} = \frac{M_{stabil}}{M_{turn\,over}} \geq 3 \quad (15)$$

$$p \leq R^{'*} \quad (16)$$

$$s_{cál} \leq s_{lím.} \quad (17)$$

expressing, respectively, the requirement of an adequate safety factor against overturning for loads on II Limit State, the condition of linearity for settlement calculation, which determines the tool for its calculation, and the limit condition for deformations (settlements).

For the structural design (ONN (NC) 2019c) the following constraints hold:

$$\tau_{pz} = \phi R_{pz} \quad (18)$$

$$V_{ul} \leq \phi V_{cl} \quad (19)$$

$$V_{ub} \leq \phi V_{cb} \quad (20)$$

$$M_{ul} \leq \phi M_{nl} \quad (21)$$

$$M_{ub} \leq \phi M_{nb} \quad (22)$$

which express, respectively, the conditions of resistance to punching, shear in both directions and positive bending in both directions of a shallow foundation slab.

For more detailed explanations and the nomenclature, we refer to the Cuban Standards (ONN (NC) 2019a, b, c, Chagoyén and Broche 2002).

*2.4 Adapted variable stiffness coefficient model for Static Soil-Structure Interaction*

By considering of SSSI in the structural optimization problem, we account for the fact that a structure, on underlying soil, deforms under loading-unloading, which produces a force redistribution in the superstructure, i.e. SSI affects the responses of the whole system during loading-unloading processes. To incorporate SSSI, the soil is assumed to be layered with constant material properties along its depth and modelled as a linearly elastic half-space, considering the limitation of depth of the compressible thickness, while the foundation is considered as a plate (shallow) footing (see Fig. 5).

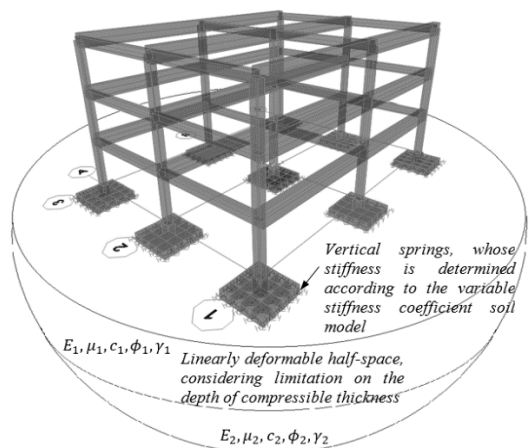

Fig. 5 Configuration for modelling the static soil-structure interaction

Several issues could also be considered in determining the soil base settlements: non-homogeneity of the geological structure of the base, non-homogeneity of the base, presence of water level, separate soil lenses and of different inclusions, the possibility of soil flooding and the change of its properties, the separate consideration of residual and elastic settlements of the base, etc.

### 2.4.1 Solution of the non-linear contact problem

The Cuban standard (ONN (NC) 2014) considers the possibility of determining the non-linear component of the soil settlements, when the acting pressure $p$ exceeds the soil linearity limit stress value $R'^*$ ($p \geq R'^*$).

In practical calculations, consideration of non-linearity was introduced in calculating settlements in collapsible soils type II (Klepikov *et al.* 1987), but the model used for this (Klepikov 1969) can be adapted and extended to other soil types and foundation typologies, such as predominantly frictional soils, where the difference between the linearity limit stress ($R'^*$) and the base load capacity ($q^*_{br}$) is larger. Therefore, the incursion in the non-linear stage of the soil behavior makes more sense in the geotechnical design of shallow foundations (Timochenko *et al.* 2015).

Approximating the relation between the acting pressure and settlement (*p* vs. *S*) for a shallow foundation resting on a soil base, by a hyperbolic equation, Klepikov (1969) defined the relation as shown in Eq. (23),

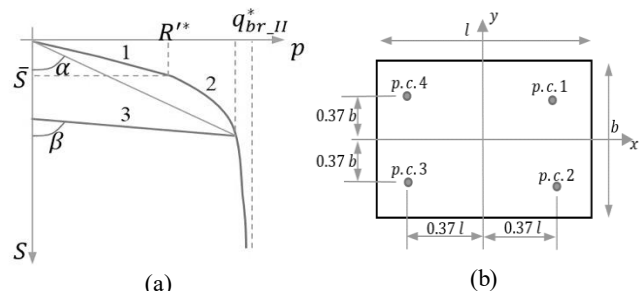

(a) (b)

Fig. 6 (a) Determination of stiffness coefficient on the non-linearly deformable base (Klepikov *et al.* 1987): diagram of calculation of the deformations of the base, (b) characteristic points in rectangular shallow foundations, according to the Cuban Standard (ONN (NC) 2004)

$$S = \frac{p \cdot \bar{S} \cdot \left(\left(\frac{q^*_{br\_II}}{R'^*}\right) - 1\right)}{q^*_{br\_II} - p} \quad (23)$$

where $p$ is the acting pressure from foundation and $\bar{S}$ is the base settlement for an acting pressure equal to the soil base linearity limit stress $R'^*$. It is noticed here, then Cuban Standard for shallow foundation design, assumes the semi-probabilistic approach of limit states design (LSD). Two Limit States (LS) are considered during foundation design: 1st LS rules bearing capacity foundation design, while 2nd. LS, rules soil behavior (linear or non-linear) during settlement calculations and methods for such determinations. In both LS, different soil properties calculation values ($\gamma, c, \phi, etc.$) are used: for the 1st. LS, design probability allowed for soil properties calculation is 95%: i. e. only 5% of soil properties values are admitted to have values below calculation values assumed for this LS calculations. During 2nd. LS calculations, design probability allowed for soil properties calculation is 85%: i. e. 15% of soil properties values are admitted to have values below calculation values assumed for this LS calculations.

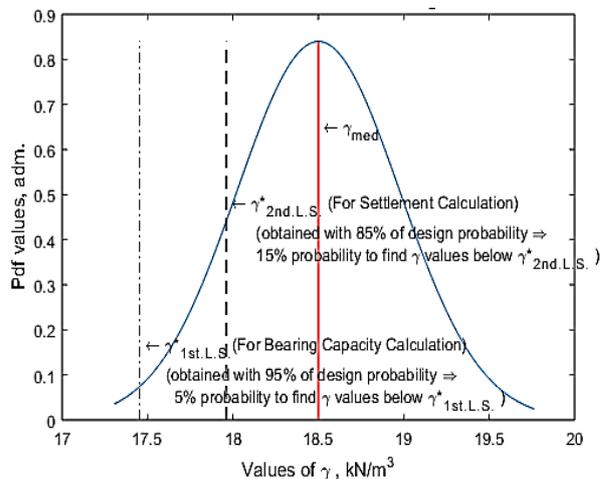

Fig. 7 On determination of calculation values for soil property $\gamma$, as per Limit State Design approach of Cuban Standard for shallow foundation design.

Fig. 7 reflects this concept for specific soil weight on its natural state, although for the rest of the soil properties, calculation values, determination is similar. The value of $R'^*$ is here determined with soil calculation properties, corresponding to 2nd. LSD, $q^*_{br\_II}$ is the base bearing capacity pressure, based on expressions from the theory of plasticity, but in this case, is also determined with soil calculation properties, corresponding to 2nd. LSD, so that, consistency between both concepts when evaluating the design safety was achieved when determining the stiffness coefficients.

The Cuban Standard recommends the Brinch-Hansen equation (Hansen 1970) for the determination of q*br_II, with c the soil cohesion, ø the soil friction angle and γ the soil specific weight as parameters.

The "pressure-settlement" curve passes through the point with coordinates $(R'^*, \bar{S})$ and unlimitedly approaches the asymptote $p = q^*_{br\_II}$ (Fig. 6(a)).

The parameters describing the non-linear diagram are determined by classical methods of soil mechanics, based on the theory of plasticity and elasticity, considering foundation dimensions and depth, and the physical and mechanical soil base properties. The relation given by Eq. 23, in addition to its simplicity, has the advantage that it allows, by means of a single curve, to describe linear and even nonlinear soil behavior, up to foundation bearing capacity, based on a set of soil calculation parameters and concepts which are familiar to designers.

The soil base "secant" stiffness coefficient $k$ can be obtained easily for the loading stage (Klepikov *et al.* 1987):

$$k = \tan\alpha = \frac{p}{S} = \frac{q^*_{br\_II} - p}{\bar{S} \cdot \left[\left(\frac{q^*_{br\_II}}{R'^*}\right) - 1\right]} \quad (24)$$

The tangent of the angle of inclination $\beta$ of the discharge line with respect to the axis of the settlements $S$ represents the stiffness coefficient of the base during discharge (Fig. 6(a)). The value is constant for the assumed deformation diagram, regardless of whether a discharge could occur at each point in the diagram.

In the foundation design process, the foundation settlements and pressures are determined, according to Cuban Standard (ONN (NC) 2014), at the so-called foundation "characteristic point", i.e. a point in which the settlement remains the same for a rigid and for a flexible foundation of the considered dimensions, so that the soil base stiffness coefficient obtained is also independent of the foundation stiffness, which is unknown during structural analysis (see Fig. 6(b)).

*2.4.2 Algorithm for SSSI modelling during optimal design of structural assemblies*

A procedure to solve the nonlinear contact problem in collapsible soils using the variable stiffness coefficient model was elaborated assuming that the dimensions of the foundation are known (Klepikov *et al.* 1987), or where they are somehow fixed by certain criteria. However, these dimensions are unknown initially.

When considering SSI in case of shallow foundations, the calculation starts by analyzing a model with classical idealized supports.

Based on the results of the first step, the geotechnical design is performed, the dimensions of the foundations are calculated and the structural design is also performed, of which only the slab thickness of the shallow foundation is required in this step. With these results, the previous model is completed with the foundation, considering its dimensions in the previous step, the stiffness coefficients that allow for an analysis with SSSI, determining the stiffness coefficient, for the i-th foundation according to the following expression:

$$k_i^{(1)} = \frac{q^{*(1)}_{br\_IIi} - p_i^{(1)}}{\bar{S}_i^{(1)} \cdot \left[\left(\frac{q^{*(1)}_{br\_IIi}}{R'^{*(1)}_i}\right) - 1\right]} \quad (25)$$

where superscript (1) represents the iteration step (first step).

The analysis of the model with consideration of SSSI is carried out next, leading to new values of internal forces. Likewise, it is possible to carry out the geotechnical and structural design again, and obtain new values for the base area and the slab thickness, with new internal forces. Based on the results of this step, $k_i^{(2)}$ is calculated, which serves as input for the solution of the contact problem in the second step of the iteration. If the foundation settlement is positive, Eq. (25) determines the stiffness coefficient. For those with negative settlements, the stiffness coefficient is zero which means that the foundation lifts from the base. For the iteration steps following the first, the coefficient can be written generally as in Eq. (26).

$$\text{If } S_i^{(d)} \geq 0 : k_i^{(d+1)} = \frac{q^{*(d)}_{br\_IIi} - p_i^{(d)}}{\bar{S}_i^{(d)} \cdot \left[\left(\frac{q^{*(d)}_{br\_IIi}}{R'^{*(d)}_i}\right) - 1\right]}$$

$$\text{If } S_i^{(d)} < 0 : k_i^{(d+1)} = 0 \quad (26)$$
$$(d = 1, 2, 3, \ldots)$$

The iterative process ends at step $h$, where the difference between the calculation results in this and the previous step $h$-1, is less than some given value, which characterizes the accuracy in the calculations (usually 5%). Results of the calculation are $k_i^{(h)}$ and the internal forces of the structure to perform the final geotechnical and structural design.

## 3. Evolutionary algorithms in structural optimization

Evolutionary strategies (EAs) are commonly used in structural optimization, due to the specific nature of the optimization problem. This is especially the case for reinforced concrete structures, where the objective function depends on many factors, resulting in many local optima, as mentioned previously (Papadrakakis *et al.* 1998, Rizzo *et al.* 2000, Serpik *et al.* 2016).

We have applied several EAs to the Ackley function (with 16 variables, trying to simulate a real problem), a benchmark function that is similar to the objective functions for the case studies in this paper (see section 4.2.1), using the toolbox of evolutionary strategies *YPEAv1.0* (Kalami Heris 2019). Biogeography-Based Optimization (BBO) outperformed the other methods available in the toolbox, especially with respect to the convergence speed, without extensive tuning of the parameters of the method.

Some authors have used this optimization method in structural optimization. Aydogdu (2017) used a version of BBO with Levy flight distribution (LFBBO) in the design optimization of cantilever retaining walls, and obtained very good results. Shallan *et al.* (2019, 2020) used BBO and GA to optimize the design of 3D and plane frame steel structures. Here, the GA performed better than BBO, but it is not clear whether parameter tuning was performed.

In general, the performance of EAs strongly depend on the chosen values for the parameters of the method, hence parameter tuning is very important, and often more important than the choice of the particular EA itself. Below we present a brief introduction to the BBO method and we explain why it can outperform other EAs. Afterwards we discuss the parameter tuning strategies that we have implemented.

### 3.1 Biogeography-Based Optimization

Biogeography studies the geographical distribution of biological organisms. Mathematical models of biogeography describe how species migrate from one habitat or island to another, how new species arise, and how species become extinct. Geographical areas that are well suited as residences for biological species are said to have a high habitat suitability index (HSI). The variables that characterize habitability are called suitability index variables (SIVs). SIVs can be considered as the independent variables of the habitat, and HSI as the dependent variable. Habitats with a high HSI have a high species emigration rate, because they host many species able to emigrate to neighboring habitats, and they have a low species immigration rate because they are already nearly saturated with species. Therefore, high HSI habitats are more static in their species distribution than low HSI habitats. Habitats with a low HSI have a high species immigration rate because of their sparse populations. This immigration may raise the HSI of the habitat, because the suitability of a habitat is proportional to its biological diversity. However, if a habitat's HSI remains low, the residing species tend to go extinct, which will further open the way for additional immigration. Due to this, low HSI habitats are more dynamic in their species distribution than high HSI habitats (Simon 2008).

The BBO method, proposed in Simon (2008), is a relatively new method based on the concepts explained above.

The correspondence between the BBO terminology and the classical EA terminology is the following: habitat, HSI and species correspond to respectively individual, fitness value and value of a variable. Hence, the number of species in each habitat is equal to the number of variables in the optimization problem. The algorithm consists of the following steps:

The algorithm starts with a random initial set of habitats with a uniform HIS distribution.

In every iteration, the emigration and immigration coefficients, denoted by respectively $\mu$ and $\lambda$, are assigned to each habitat. Solutions or habitats with a high HSI receive high values of $\mu$ and low values of $\lambda$, and vice versa.

The algorithm processes the habitats in order of decreasing HSI, using the parameters $\mu$, $\lambda$ and mutation probability as follows.

- Within each habitat, for each species the possibility to carry out the migration process is analyzed: each species is checked based on the habitat's immigration coefficient $\lambda$. Therefore, the species of the best habitats have little chance of entering this process, while this chance increases when considering habitats with a higher $\lambda$.
- Once a species enters the migration process, another species from other habitat is selected using *roulette wheel selection* (to select the habitat) based on $\mu$ to immigrate to the habitat being worked on.
- Once the species are selected, immigration starts, which is not the substitution of one by the other, but a combination of both, performed as:

$$NewSpecies^i_k = Species^i_k + \alpha(Species^j_k - Species^i_k) \quad (27)$$

where $Species^i_k$, i.e. the *k*-th species of habitat *i*, is the species being analyzed and $Species^j_k$, i.e. the *k*-th species of habitat *j*, is the species selected to immigrate, and $\alpha$ is the acceleration coefficient (0.9 per default).

- In addition, species can mutate with a certain probability according to:

$$NewSpecies^i_k = NewSpecies^i_k + \sigma N(0,1) \quad (28)$$

where $\sigma$ is the mutation step size (0.05 per default), $N(0,1)$ is a random number with mean 0 and standard deviation 1. After every iteration, $\sigma$ decreases, modified by the *mutation step size damping* (0.99 per default).

Once the entire population is analyzed, the new one is formed by selecting the best habitats of the previous population (before being transformed) and the best of the new population. The fraction of the previous population that survives is denoted by *KeepRate*. This is similar to elitism used in GA.

This iterative process ends when a stop criterion is satisfied. In section 4.3 we study the effect of the parameters population size (*PopSize*), keep rate (*KeepRate*), acceleration coefficient (*Alpha*) and mutation probability (*MutProb*).

### 4. Parameter tuning

Parameter tuning can improve the performance of an optimization algorithm (Eiben and Jelasity 2002, Smit and Eiben 2010a, Huang *et al.* 2019), but can be tedious and difficult to implement. A very useful tool to determine suitable values for the parameters is the study of utility landscapes, which are abstract landscapes over the space of the parameter vectors of an Evolutionary Algorithm, where the height corresponds to the utility, based on an appropriate notion of EA performance (Smit and Eiben 2010a).

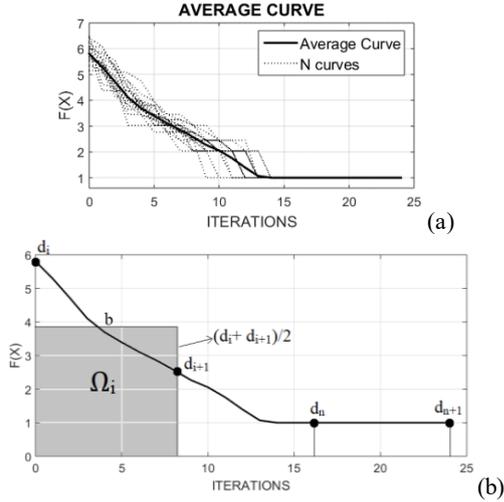

Fig. 8 Utility function: (a) the average performance curve of the N curves, (b) visualization of utilities A and B

According to Eiben and Smit (2011), four groups of methods to perform parameter tuning can be distinguished: sampling methods, model-based methods, screening methods and meta-evolutionary algorithms. Taking this classification as a reference, and based on our needs, we use a methodology based on a combination of sampling and screening methods, by generating utility landscapes for each possible combination of the parameters analyzed.

### 4.1 Utility metrics

To measure the performance of an EA one must take into account the quality of the solution and the computational cost of the algorithm. Most, if not all, performance metrics used in evolutionary computation are based on variations and combinations of these two. Solution quality can be naturally measured by using the fitness function (Eiben and Smit 2011).

In addition, due to the stochastic nature of EAs, it is necessary to perform a certain number of tests to obtain reliable performance measures. The following performance measures, also called utilities, can be used (Eiben and Smit 2011): mean best fitness (MBF), average number of evaluations to solution (AES) and success rate (SR).

In Eiben and Smit (2011) a summary of the most used utilities is given, stating that these measures are not always appropriate. For example, in case of large variance in the performance results of the EA, the use of the mean (and standard deviation) may not be significant and the use of median or best fitness may be preferable (Bartz-Beielstein 2005). Other procedures could also be used such as visualizations using boxplots (Dillen *et al.* 2018) and plots of the Empirical Cumulative Distribution Function (ECDF) (Chiarandini *et al.* 2007). The performance metrics determine the utility landscape and, therefore, the optimal parameter vector. Hence, the final results of the EA may vary substantially depending on the utility used. Therefore, care must be taken when defining it (Smit and Eiben 2010b).

We assume that the EA is run *N* times with a given fixed parameter vector to reduce the stochastic effects. The average performance curve (average fitness in function of the iteration number), denoted as $f_a(x)$, is obtained by averaging the performance curves of the *N* runs of the EA (see Fig. 8(a)). This curve gives more information about performance than simple performance measures, such as MBF or AES, and allows an easy comparison between two or more procedures.

We present three ways to measure utility, with A and B being the basis for creating utility C, which is what we propose for this type of study.

Utility A

We use MBF as utility A, which can also be written as in Eq. (29).

$$F_A = d_{n+1} \qquad (29)$$

Where $d_{n+1}$ is the last point of the average performance curve (see Fig. 8(b)). However, for relatively simple problems, where the EA finds the same global or local optimum for various parameter vectors, it is impossible to select the best parameter vector using only this utility, especially in an automated procedure.

Utility B

Utility B takes into account the convergence speed of the algorithm, by using the area under the average performance curve. By splitting the number of iterations in n equal intervals of length b and by defining $d_i$, $i = 1,…, n+1$ as the values of the average fitness at the end points of the intervals (Fig. 8(b)) we approximate the area under the average performance curve by Eq. (30).

$$B = \sum_{i=1}^{n}\left(\frac{d_i + d_{i+1}}{2}\right)b \qquad (30)$$

Note that this is equivalent to using the trapezoidal quadrature formula to approximate the integral. In order to make this utility "compatible" with utility A, Eq. (30) is rescaled, yielding as in Eq. (31).

$$F_B = \frac{\sum_{i=1}^{n}\left(\frac{d_i + d_{i+1}}{2}\right)}{n} \qquad (31)$$

For the average performance curves in this paper, $n = 14$ intervals are sufficient to obtain a good approximation of the area under the curve.

However, this utility does not sufficiently take into account MBF, since it is possible that an EA that converges fast to a rather high MBF has a lower $F_B$ than an EA that converges slowly to a substantially better MBF.

Utility C

Utility C is a weighted linear combination of utilities A and B, with a higher weight on utility A, to give higher weight on the solution quality than on the convergence speed, hence:

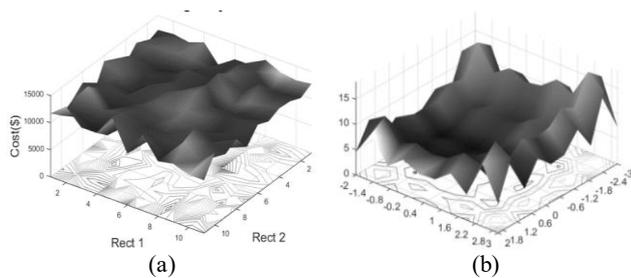

(a)          (b)

Fig. 9 Use of analytical functions to generate response surfaces similar to those generated in real problems: (a) real response surface taking into account rectangularity of foundations, (b) discretization of Ackley function to simulate the response surface in (a)

$$F_C = \frac{Z_1 * F_A + F_B}{1 + Z_1} \quad (32)$$

with $Z_1 \geq 1$, the weight that we give to utility A, i.e. MBF. In this paper we use $Z_1 = 4$.

### 4.2 Cost reductions

As mentioned, in structural optimization processes using design software for modeling and analysis, the evaluation of the objective function is computationally expensive. Hence, parameter tuning, requiring many runs of the EA and thus many function evaluations, is a very costly procedure. Therefore, obtaining a simple, but efficient methodology is essential. We discuss several alternatives to reduce the computational cost.

#### 4.2.1 Use of analytical functions

We can replace the objective function used in a real optimization problem by a simple analytical mathematical function in order to simulate the main features of the real problem. Some benchmark functions (Suganthan *et al.* 2005) can be used for this purpose. Since these functions are continuous, we discretize them and we define suitable intervals for each variable. For example, the 2D Ackley function behaves quite similar as the cost of RC structures in function of the rectangularity of foundations (see Fig. 9).

Another alternative, also used in this investigation, is to create a database containing objective function values for a large number of points in the parameter space, computed using the calculation engine. During a run of the EA, the required values of the objective function can be obtained by interpolation of the values stored in the database, or can be computed exactly and stored in the database, which then grows during the run. Especially for parameter tuning this can lead to huge savings in computational cost.

### 4.3 Utility landscape analysis

Once the utility is defined, we can compute and analyze utility landscapes. The stochastic nature of the EA must be taken into account, as well as the parameters to be tuned and their possible values, in order to perform reliable parameter tuning. In case of BBO, we consider four parameters: *PopSize, KeepRate, Alpha* and *MutProb* (see section 3). Considering simultaneously the four parameters requires the computation of the utility for many points in the four-dimensional parameter space.

Considering less parameters simultaneously yields a low-dimensional utility landscape. Less points on the landscape must be computed, but for each point, more runs of the EA are needed to get reliable results due to the stochastic behavior in case random values for the other parameters are selected. To evaluate the effect of varying one specific parameter on the utility, a 1D utility landscape can be computed: to determine a point on this landscape, the value of the selected parameter is fixed, while the other parameters take random values. The resulting 1D utility landscapes are easy to interpret, e.g. by using box plots. Another efficient way to check the results would be using the performance curve itself.

A simple experiment using four parameters of BBO for optimizing the Ackley function (see further) indicates that to compute a 1D utility landscape, by considering only one parameter, while the other three parameters take random values, the number of runs of the EA needed to get reliable results for utility C is 5 times higher than when 2D landscapes are computed, by considering two parameters and 15 times higher than when the 4D landscape is compute.

To tune the parameters of the BBO method, we computed 2D utility landscapes, by considering each time two parameters, while the values of the other parameters were randomly selected. These 2D utility landscapes, computed for the various utilities, allow to easily verifying the influence of the combination of two parameters.

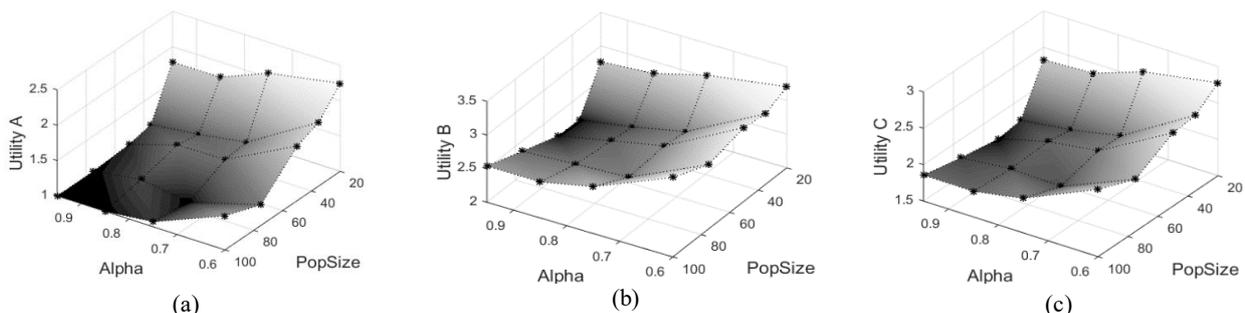

(a)          (b)          (c)

Fig. 10 2D utility landscapes for BBO applied to the Ackley function with parameters *Alpha* and *PopSize*. (a), (b) and (c) refer to the respective utility

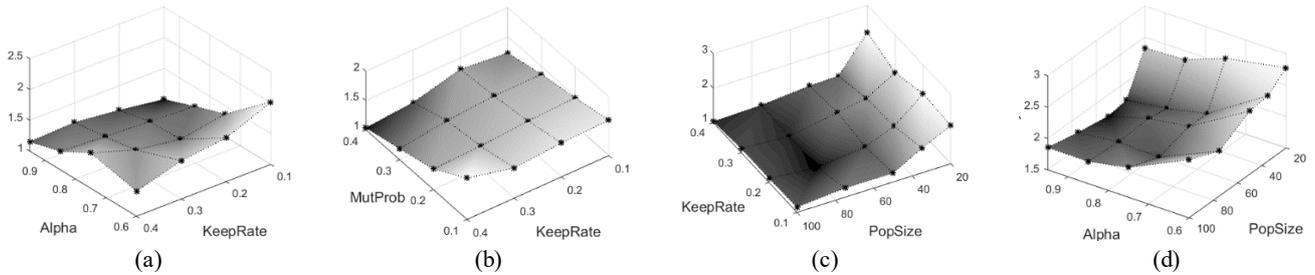

Fig. 11 2D utility landscapes for BBO used to optimize the Ackley function, using utility C

Fig. 10 shows the 2D utility landscapes for BBO with parameters *Alpha* and *PopSize*, using the three utilities A, B and C. For the parameters *Alpha* and *PopSize*, respectively 4 and 5 values were selected, leading to 20 parameter combinations, for which the utility is computed. For each of these parameter combinations, 30 runs of the BBO algorithm are executed, with random values for the parameters *KeepRate* and *MutProb*. These landscapes provide useful information, such as the effect of the utility used. Fig. 10(a) shows that 100% SR is obtained for five combinations of parameter values for (*PopSize, Alpha*), i.e. (60, 0.95), (80, 0.70), (80, 0.95), (100, 0.95) and (100, 0.95). As already explained in the previous section, the optimal parameter combination based on utility B (*PopSize, Alpha*) = (40, 0.95) does not guarantee an optimal SR, while utility C selects, among the parameter combinations leading to an optimal SR, the combination for which convergence is fastest, i.e. (*PopSize, Alpha*) = (60, 0.95).

In addition, the influence of each parameter separately on the method's performance can be extracted from the utility landscapes, and this information can be used to reduce the process cost. Once we observe that certain parameter has only limited influence on the utility of an EA, we can remove this parameter from the tuning process. In Figs. 11(a)-(b)-(c) it is evident that the parameter *KeepRate* has little influence on utility C, so this parameter can be neglected in the tuning process.

Additionally, the range of parameter values considered in the tuning process can be restricted. For example, the utility is poor for small values of *PopSize* (Fig. 11(c)-(d)) and for small values of *Alpha* (Fig. 11(a)-(d)).

### 4.3 Parameter tuning results

We first present results obtained by the BBO optimization method for the six case studies described in section 1 and we discuss the parameter tuning. Afterwards we apply the BBO method to a real test case, i.e. the Sindico House with the tuned parameter values.

#### 4.4.1 Simple case studies

Based on the information of previous section (using Ackley function), and taking into account the results obtained when starting the study of the case studies, we performed the following parameter tuning.

We computed 2D utility landscapes for case study 1, as described in section 4, using several values for the four parameters of the BBO method. The results shown in Fig. 12 indicate that the value of *KeepRate* indeed has not much influence on the optimization results and we decided to set *KeepRate* = 0.40. In many cases, the best result was obtained for large values of the other parameters. Therefore, we selected the following values of the parameters during the parameter tuning discussed below: *PopSize* = [60 80 100 120 140], *Alpha* = [0.90 0.95 0.99] and *MutProb* = [0.30 0.40 0.50].

We only show the most significant utility landscapes for each case study.

From Fig. 13 and other utility landscapes (not shown), we derived the best values for each parameter. In addition, for each test case we executed the BBO optimization with the best parameter values, obtained for the other test cases, with the objective to investigate whether the BBO method is a good generalist. The results are summarized in Table 2.

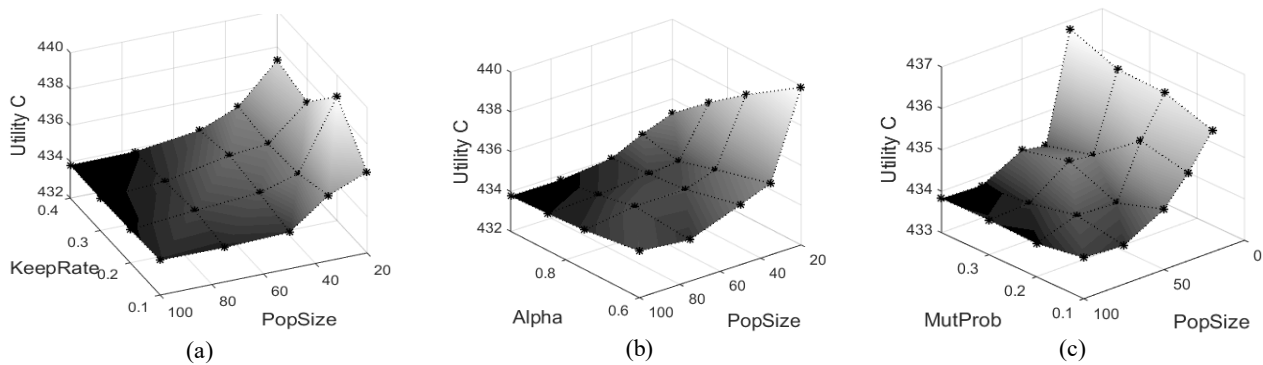

Fig. 12 Utility landscapes for case study 1, which prove some partial conclusions previously obtained

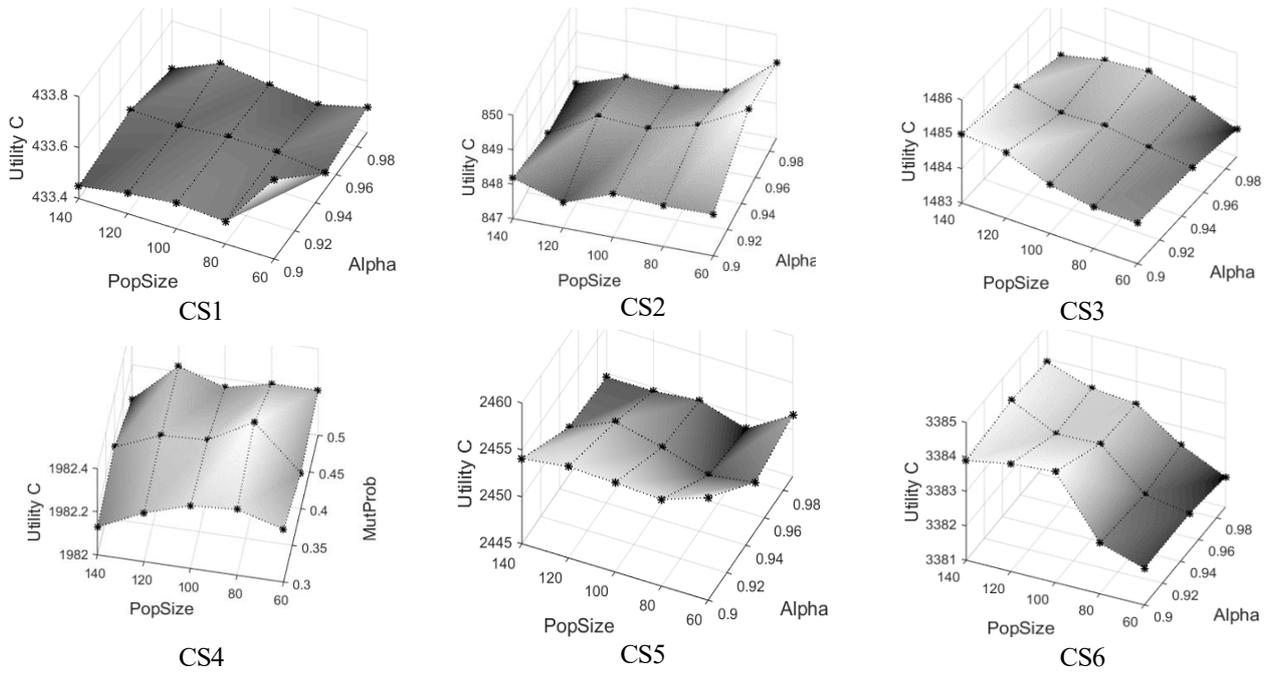

Fig. 13 Most representative utility landscapes for the six case studies, using Utility C

Table 2 Best values for *PopSize, Alpha* and *MutProb* obtained by tuning the parameters for each case study and the values of Utility C for each set of tuned values for all case studies. The scaled utility *ScUt* is mentioned between brackets.

| Best values for each problem (*PopSize, Alpha, MutProb*) | C utility (scaled utility *ScUt*) for each case study | | | | | |
|---|---|---|---|---|---|---|
| | CS 1 | CS 2 | CS 3 | CS 4 | CS 5 | CS6 |
| (140,0.99,0.5)* | 433.43 (99.98) | 847.49 (99.92) | 1484.59 (99.89) | 1982.12 (99.98) | 2449.10 (99.86) | 3384.30 (99.83) |
| CS3 (60,0.99,0.4) | 433.51 (99.96) | 848.65 (99.87) | 1483.96 (99.93) | 1982.24 (99.98) | 2449.37 (99.85) | 3382.40 (99.89) |
| CS5 (80,0.99,0.5) | 433.50 (99.97) | 847.80 (99.88) | 1484.12 (99.92) | 1982.17 (99.98) | 2448.40 (99.89) | 3382.22 (99.89) |
| CS6 (60,0.99,0.5) | 433.48 (99.97) | 848.51 (99.84) | 1483.96 (99.93) | 1982.28 (99.98) | 2449.60 (99.84) | 3381.70 (99.90) |

* Values of the tuned parameters were the same for CS1, CS2 and CS4

In addition, in this table we show (enclosed in parentheses) a scaled utility (*ScUt*). It allows comparing the utility C value with respect to the minimum single value obtained in the whole tuning process (or global optimum if the function is known), which represents the best possible value (*BestUt*) that Utility C can take. This scaled utility represents how close the values of Utility C (*CUtil*) and the best utility value are (in percent), and it is computed as in Eq. 33.

$$ScUt = \left[1 - \left[\frac{(CUtil - BestUt)}{CUtil}\right]\right] * 100 \qquad (33)$$

If *ScUt* = 100%, the best value was obtained with Utility C.

The data in Table 2 does not allow to draw conclusions about the best value for *PopSize* when solving problems of varying complexity. For this reason, based on a proposal in Eiben and Smit (2011), the scaled utility is visualized in Fig. 14 for several values of *PopSize* and for each of the six cases. Clearly *PopSize* = 80 is a good choice, since it provides good results for the six problems. Other values that, according to Table 2, are better for certain cases, are not suitable for other cases (e.g. 60 is not optimal for case study 5, and 140 is not optimal for case studies 5 and 6).

On the other hand, case studies 1 and 4 were the easiest to optimize. This indicates that the BBO method has more difficulties with problems in which the variables can take many values. Hence, the number of variables does not seem to be as important as the number of values they can take.

### 4.4.2 Síndico House project

By studying this much more complex structure, we aim to validate the conclusions w.r.t. to the optimization of the structure and the parameter tuning from the simple case studies.

We consider the Síndico House structure without SSSI (Problem 1 with 12 variables) and with SSSI (Problem 2 with 16 variables). The scaled utility for various values of *PopSize* in the BBO optimization procedure for both problems, see Fig. 15, clearly indicates that *PopSize* = 80 is the best choice.

### 5. Structural results and discussions

We now discuss the structural results of the optimization processes. Selecting span/height ratio (*L/h*) and reinforcement ratio ($\rho$) in beams, and rectangularity for columns and foundations allow comparisons with other studies in the literature.

Table 3 Summary of the most significant structural results for the six case studies

|  |  | Results | CS1 | CS2 | CS3 | CS4 | CS5 | CS6 |
|---|---|---|---|---|---|---|---|---|
| Beams | | L/h | 8 | 8.4 | 17 | 17 | 17 | 17 |
|  | $\rho$ (%) | Bottom | 0.60 | 0.72 | 0.71 / 0.57 | 1.03 | 0.76 / 0.57 | 1.03 |
|  |  | Top | 0.16 | 0.17 | 0.88 / 0.88 | 1.59 | 1.01 / 1.03 | 1.77 |
| Columns Rectangularity | | Int | - | - | - | 1.30 | - | 1.30 |
|  | | Ext | 1.40 | 1.40 | 1.15 | 1.00 | 1.15 | 1.00 |
| Foundations Rectangularity | | Int | - | - | - | - | - | 1.50 |
|  | | Ext | - | 1.20 | - | - | 0.50 | 1.75 |

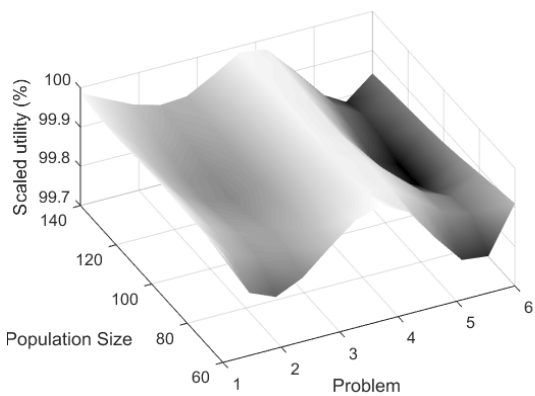

Fig. 14 Scaled utility landscape of *PopSize* parameter for each problem

### 5.1 Simple case studies

Table 3 shows certain trends. In the case of beams, for case studies 1 and 2, the optimal heights are larger than for the other cases. This can be understood because in simple structures beams play a fundamental role as a stiffener or flexure regulator in the rest of the structure, increasing the need to reinforce the columns. The design of columns is especially susceptible to bending. In more complex structures this influence is lost and the optimal height are smaller.

The case of $\rho$ is the most relevant aspect in these optimization procedures. It is associated with the inclusion of SSSI. While beams with identical sections are obtained when SSSI is included, differences in $\rho$ of up to 0.18 % are obtained. That represents 1.6 cm$^2$ or almost 1 bar of ø 16. This difference mainly lies in top reinforcement at supports, since differential settlements induce major changes in design forces in that zone. The results may vary depending on the geometric characteristics of the structure, soil properties, etc.

The columns with the longest sides are in the direction of greatest flexion due to gravitational loads. For cases 4 and 6, they are also supporting the stiffness in the most critical direction due to wind load. Foundations are rectangular conditioned by the critical direction of the wind load.

### 5.2 Síndico House project

We now consider a structure of greater structural complexity with an asymmetric floor plan, i.e. Síndico House, see Fig. 2(b). The soil under the structure is predominantly frictional. Consequently, there is a combination of circumstances causing it to be a very sensitive case study with respect to the objectives being analyzed.

Our discussion of the results for Síndico House structure is based on similar aspects as for the simple case studies. Table 4 shows the results of four design processes. The first two do not include SSSI in modeling. We further compare the use of a simple design process, without optimization, and the optimized design processes using the BBO procedure. To perform the simple design, the criteria adopted by the project designers are used: *L/h* ratios of 14 in beams, columns with square cross section, foundations with square footings and concrete with f'c=25 MPa for all the elements.

With the inclusion of SSSI, the direct costs are higher: +23 % in the superstructure and +20 % in total. This does not mean that not including SSSI is more rational, but we are performing designs with significant difference from a safe and efficient design. In addition, the optimal heights of the beams decrease in the optimized design (it does not happen in simple case studies). When making an analysis of ρ, similar to what was done in simple models (use of equal cross sections), we can observe differences of up to 0.63 % (2 bars of ø 19).

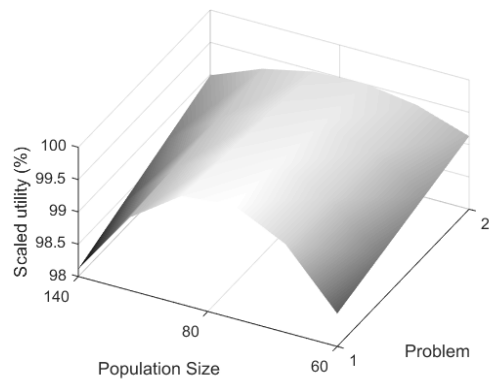

Fig. 15 Scaled utility landscape of *PopSize* parameter for problems 1 and 2

Table 4. Summary of the most significant structural results for Síndico House project

| | | | No SSSI Simple* | No SSSI BBO | SSSI Simple* | SSSI BBO |
|---|---|---|---|---|---|---|
| Beams** | L/h | Group 1 | 14 | 20 | 14 | 15.50 |
| | | Group 2 | 12 | 20 | 12 | 15.00 |
| | ρ (%) Bott | Group 1 | 0.68 | 1.30 | 0.68 | 0.80 |
| | | Group 2 | 0.62 | 1.03 | 0.68 | 0.99 |
| | Top | Group 1 | 1.23 | 2.59 | 1.82 | 2.02 |
| | | Group 2 | 0.79 | 2.12 | 1.82 | 1.79 |
| | f´c (MPa) | | 25 | 20 | 25 | 30 |
| Columns | Rectang | Interiors | 1 | 1.15 | 1 | 0.86 |
| | | Exteriors | 1 | 1 | 1 | 1.67 |
| | | Corners | 1 | 0.86 | 1 | 0.67 |
| | f´c (MPa) | | 25 | 35 | 25 | 35 |
| Foundations | Rectang | Interiors | 1 † | 1 † | 1 | 0.50 |
| | | Exteriors | 1 † | 1 † | 1 | 1.25 |
| | | Corners | 1 † | 1 † | 1 | 1.50 |
| | f´c (MPa) | | 25 † | 25 † | 25 | 20 |
| Direct cost ($) | Superstructure | Beams | 11765.90 | 9251.31 | 11808.17 | 10803.51 |
| | | Columns | 5019.50 | 5717.19 | 10079.88 | 5819.37 |
| | | TOTAL | 16785.40 | 14968.50 | 21888.05 | 16622.88 |
| | Foundations | | 5407.64 | 5738.93 | 6031.37 | 5295.65 |
| | TOTAL | | 22193.04 | 18859.03 | 27919.42 | 21918.53 |

\* Simple design process using the designers' criteria

\** Two most representative design groups are represented (L=7 y 6 m respectively)

† Foundations are not taken into account in modeling without SSSI, but they are designed, and their cost is calculated

In Fig. 16(a), the cost of the columns in the third variant is remarkable. This result is because columns in this asymmetric structure (with SSSI also included) hold a considerable bending moment, so the use of columns with square cross section leads to an increase of the direct cost. The previous observation is even more visible in Fig. 16(b). Compared to the optimized design (SSSI BBO) it can be seen how searching for a rational configuration of the columns' sections, the implemented optimal design algorithm finds the column shape capable to face bending moment more efficiently, reducing the reinforcement cost by 51%, even when the concrete cost is reduced by 25%.

Overall, a substantial direct cost reduction of 21% is achieved. This is due to a combination of an asymmetric structure with a predominantly frictional soil with a large bearing capacity. This causes a non-linear soil behavior (curve 2 in Fig. 6(a)), under load cases for the II Limit State. In this situation, small differences between contiguous foundations (acting forces change for elements with the same sizes, for example) result in differential settlements, causing superstructure design forces redistribution. For this reason, some results may be surprising, for instance, columns with high rectangularity,

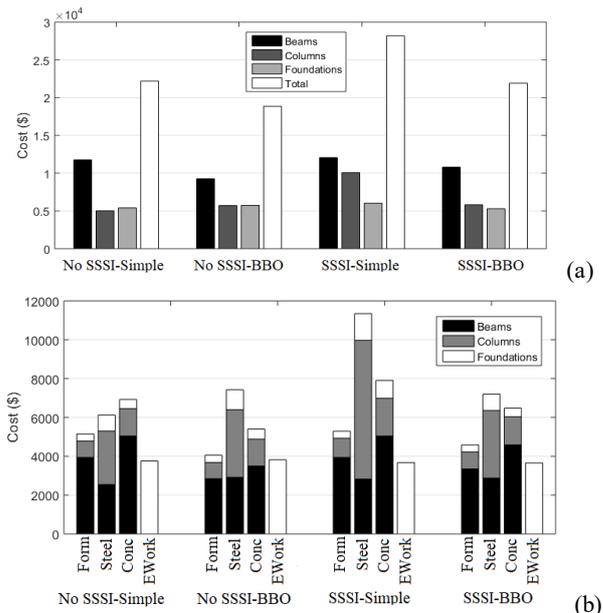

Fig. 16 Graphic distribution of direct costs: a) cost distribution by elements in each variant, b) breakdown of total direct cost in components (formwork, steel, concrete and earthwork) for each variant

beams with high-quality concrete (when the elements that work in flexion do not need it) to guarantee more vertical rigidity of the structure ($E = 4700\sqrt{f´c}$) and significant differences (in direct cost) when not including SSSI or when certain criteria are used in a simple design instead of an optimized design.

## 6 Conclusions

In this paper, we propose a methodology to optimize the design of three-dimensional concrete frame structures, including static soil-structure interaction, using an evolutionary algorithm, namely Biogeography-Based Optimization (BBO).

• Extensive parameter tuning of the BBO method has been performed. To limit the cost of parameter tuning, we initially used a surrogate analytical function and afterwards a database with objective function values for the real models.

• We have proposed several utilities to assess the performance of the BBO method. The optimal values of the parameters of the BBO method for the optimization of models of increasing complexity have been found by analyzing utility landscapes. We have observed that the optimal values are not very sensitive to variations in the models.

• BBO, with parameters tuned for optimizing rather simple structures, has been used for the optimization of the design process of an actual building with an asymmetric floor plan. The results indicate that using the BBO optimization algorithm, instead of designers' common practices, can result in substantial savings.

• For beams, higher reinforcement ratios have been obtained, when compared to a conventional design, while for columns rectangular cross sections have been selected, as opposed to the more common square shape in response to the dominant direction of bending. Shallow foundations also get their more rational configurations with rectangular footings.

• Further, the importance of taking into account static soil-structure interaction (SSSI) has been verified. In case SSSI is taken into account, non-negligible differences in the reinforcement ratios of the beams have been obtained, and the direct costs of the superstructure increase substantially as well, when compared with a conventional design. This obviously does not mean that omitting SSSI leads to a more rational design, but implies that we are using a significant percentage of less material than what is needed for an efficient design.

## Acknowledgments

We acknowledge the financial support of VLIR-UOS via the projects "Computational Techniques for Engineering Applications" (ZEIN2012Z106) and "Vibration Assessment of Civil Engineering Structures" (ZEIN2016PR419), allowing the first author to follow postgraduate courses, to perform a study visit at KU Leuven and to use the HPC infrastructure of the VSC (Flemish Supercomputer Centre).